%% file: CoBRA.tex
\documentclass[letterpaper, 10 pt, conference]{ieeeconf}  

\IEEEoverridecommandlockouts                              

\usepackage{graphicx}
\usepackage{amscd}
\usepackage{amssymb}
\usepackage{epsf}
\usepackage{pgf}
\usepackage{url}
\usepackage{flushend} 

\usepackage{amsmath,amsfonts}
\usepackage[T1]{fontenc}  
\usepackage{psfrag}
\usepackage{algorithm}  
\usepackage{algpseudocode}
\usepackage{pifont}
\usepackage{cite}
\usepackage{hyperref}
\usepackage{booktabs}
\usepackage[amsmath,hyperref,thmmarks]{ntheorem}  
\usepackage{pgfgantt}
\usepackage{import}
\usepackage{listings} 
\usepackage{bm} 
\usepackage{bbm} 
\usepackage{footnote} 
\usepackage{dirtree} 
\usepackage[separate-uncertainty = true]{siunitx}  
\usepackage{seqsplit} 
\usepackage{microtype}  

\theorembodyfont{\normalfont}
\theoremstyle{plain}
\theoremsymbol{\ensuremath{\square}}
\theoremheaderfont{\bf}
\theoremseparator{:}

\DeclareMathOperator*{\argmin}{arg\,min}
\DeclareMathOperator{\sign}{sign}

\newcommand{\eef}{\mathrm{eef}}
\newcommand{\desired}{\mathrm{d}}
\newcommand{\external}{\mathrm{ext}}
\newcommand{\dynamics}{\mathrm{dyn}}
\newcommand{\given}{\mathrm{given}}
\newcommand{\final}{\mathrm{f}}
\newcommand{\duration}{\mathrm{d}}
\newcommand{\com}{\mathrm{CoM}}  

\newcommand{\at}{\mathrm{at}}
\newcommand{\reach}{\mathrm{reach}}
\newcommand{\followed}{\mathrm{followed}}
\newcommand{\returnto}{\mathrm{returnTo}}
\newcommand{\pause}{\mathrm{pause}}
\newcommand{\leftw}{\mathrm{left}}

\newcommand{\reals}{\mathbb{R}}  
\newcommand{\naturals}{\mathbb{N}}  
\newcommand{\constraints}{\mathcal{C}}  
\newcommand{\G}{\mathcal{G}}  
\newcommand{\workspace}{\mathcal{W}}  
\newcommand{\occupancy}{\mathcal{O}}  
\newcommand{\powerset}{\mathcal{P}}  
\newcommand{\links}{\mathcal{L}}
\newcommand{\SEThree}{\mathbb{SE}(3)}
\newcommand{\SOThree}{\mathbb{SO}(3)}

\newcommand{\benchmark}{B}
\newcommand{\modules}{\mathcal{R}}  
\newcommand{\costid}{C}
\newcommand{\costfun}[1]{J_{#1}}
\newcommand{\task}{\Theta}
\newcommand{\goal}[1][]{
    \ifthenelse{\equal{#1}{}}{G}{G_{#1}}
}
\newcommand{\constraint}[1][]{
    \ifthenelse{\equal{#1}{}}{c}{c_{#1}}
}

\newcommand{\T}{\mathbf{T}}  
\newcommand{\Teff}{\T_{\eef}}
\newcommand{\Tdes}{\T_{\desired}}
\newcommand{\B}{\mathbf{B}}  
\newcommand{\Ident}{\mathbf{I}}
\newcommand{\Proj}{S}
\newcommand{\Diff}[2]{\Delta_\Proj(#1, #2)}  
\newcommand{\rotM}{\mathbf{R}}

\newcommand{\Jposv}{\mathbf{q}}  
\newcommand{\Jvelv}{\dot{\mathbf{q}}}  
\newcommand{\Jaccv}{\ddot{\mathbf{q}}}  
\newcommand{\Jtorqv}{\bm{\tau}}  
\newcommand{\Jmotionv}{\mathbf{z}}
\newcommand{\Jpos}{q}  
\newcommand{\Jvel}{\dot{q}}  
\newcommand{\Jacc}{\ddot{q}}  
\newcommand{\Jtorq}{\tau}  

\newcommand{\Transv}{\mathbf{t}}
\newcommand{\Pointv}{\mathbf{p}}
\newcommand{\Fext}{\mathbf{f}_{\external}}
\newcommand{\Zerov}{\mathbf{0}}
\newcommand{\Tolboundv}{\Gamma}
\newcommand{\Tolbound}{\gamma}
\newcommand{\vtrans}{\mathbf{v}}
\newcommand{\vrot}{\bm{\omega}}
\newcommand{\velocity}{v}  

\newcommand{\timenow}{t}
\newcommand{\timezero}{0}
\newcommand{\timeend}{t_{\mathrm{N}}}
\newcommand{\timeany}{t'}  
\newcommand{\timefinal}[1]{t_{\final, #1}}
\newcommand{\timeduration}[1][]{
    \ifthenelse{\equal{#1}{}}{t_{\duration}}{t_{\duration}( #1 )}
}  

\newcommand{\body}[1]{b_{#1}}
\newcommand{\joint}[1]{j_{#1}}
\newcommand{\link}{l}
\newcommand{\assembly}{M}
\newcommand{\module}[1]{m_{#1}}
\newcommand{\connector}{\sigma}  
\newcommand{\childBody}{\chi}
\newcommand{\parentBody}{p}
\newcommand{\bodyOcc}[1]{\occupancy_{\body{#1}}}
\newcommand{\gearRatio}{k}
\newcommand{\rotorInertia}{I_{\mathrm{m}}}

\newcommand{\mechE}{J_{\mathrm{mechE}}}  
\newcommand{\Whit}{J_{\mathrm{Whit2}}}  
\newcommand{\Liu}{J_{\mathrm{Liu2}}}  
\newcommand{\numJ}{J_{\mathrm{numJ}}}  
\newcommand{\massJ}{J_{\mathrm{m}}}  
\newcommand{\timeJ}{J_{\mathrm{T}}}  
\newcommand{\fviscous}{f_{\mathrm{v}}}  
\newcommand{\fcoulomb}{f_{\mathrm{c}}}  

\newcommand\copyrighttext{%
  \footnotesize \textcopyright 2024 IEEE. Personal use of this material is permitted. Permission from IEEE must be obtained for all other uses, in any current or future media, including reprinting/republishing this material for advertising or promotional purposes, creating new collective works, for resale or redistribution to servers or lists, or reuse of any copyrighted component of this work in other works.
  DOI: TBA (Accepted at IEEE ICRA'24)}
\newcommand\copyrightnotice{%
\begin{tikzpicture}[remember picture,overlay]
\node[anchor=south,yshift=10pt] at (current page.south) {\fbox{\parbox{\dimexpr\textwidth-\fboxsep-\fboxrule\relax}{\copyrighttext}}};
\end{tikzpicture}%
}

\usepackage{mathtools}

\title{\LARGE \bf
CoBRA: A Composable Benchmark for Robotics Applications}

\author{Matthias Mayer, Jonathan K\"ulz, and Matthias Althoff
\thanks{All authors are with 
the Technical University of Munich, TUM School of Computation, Information and Technology,
Chair of Robotics, Artificial Intelligence and Real-time Systems, Boltzmannstra{\ss}e 3, 85748, 
Garching, Germany.
{\tt\small $\{$\href{mailto:matthias.mayer@tum.de}{matthias.mayer}, \href{mailto:jonathan.kuelz@tum.de}{jonathan.kuelz}, \href{mailto:althoff@tum.de}{althoff}$\}$@tum.de}}
}

\begin{document}

\bstctlcite{IEEEexample:BSTcontrol}  

\maketitle
\thispagestyle{empty}
\pagestyle{empty}

\urlstyle{same} 
\begin{abstract}
Selecting an optimal robot, its base pose, and trajectory for a given task is currently mainly done by human expertise or trial and error. 
To evaluate automatic approaches to this combined optimization problem, we introduce a benchmark suite encompassing a unified format for robots, environments, and task descriptions. 
Our benchmark suite is especially useful for modular robots, where the multitude of robots that can be assembled creates a host of additional parameters to optimize.
We include tasks such as machine tending and welding in synthetic environments and 3D scans of real-world machine shops.
All benchmarks are accessible through \href{https://cobra.cps.cit.tum.de}{cobra.cps.cit.tum.de}, a platform to conveniently share, reference, and compare tasks, robot models, and solutions. 
\end{abstract}
\urlstyle{tt}

\copyrightnotice

\section{Introduction}
\label{sec:introduction}

Scientific research benefits when results are reproducible and easily comparable to alternative solutions. 
For instance, in computer science and robotics, computer vision benchmarks like ImageNet \cite{Imagenet_ILSVRC15} or MS-COCO \cite{MSCOCO14} have brought tremendous progress. 
One key feature is that they broke down visual perception into tasks of varying difficulty, from single, cropped frame labeling to detecting multiple objects. 
These benchmarks certainly coincided with the resurgence of (deep) learning and possibly enabled it in the first place \cite{MSCOCO14}.
An area in robotics where multiple benchmarks exist is grasping and/or bin picking \cite{ColumbiaGrasp09, OpenGrasp11, mahler2019learning}; more are discussed in \cite[Tab.~1]{Chamzas2022}. 
Especially Dex-Net \cite{mahler2019learning} co-develops both novel solutions for grasp planning and improving them through publishing data sets for training and evaluation.

\begin{figure}
    \vspace*{1.7mm}  
    \centering
    \resizebox{\columnwidth}{!}{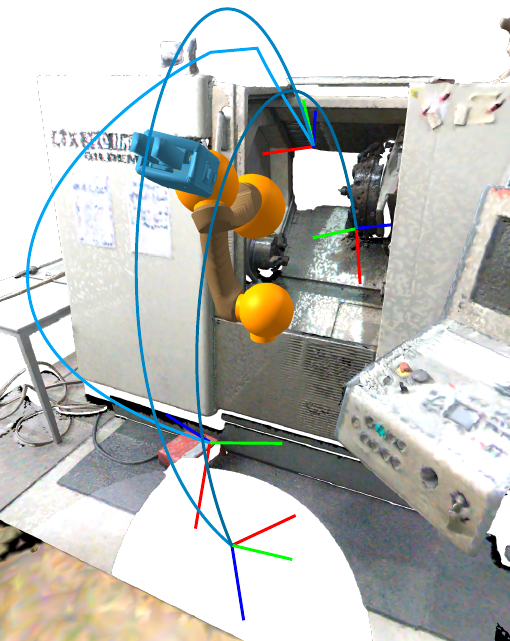}
    \caption{
    	A robot solving the machine tending task Liu2020/Case2b/Schunk\_ IWB\_CTX\_300\_linear\_0 minimizing mechanical energy. 
    	The robot is assembled with the module set $\modules = \mathtt{IMPROV}$ in the order $\assembly = [1, 21, 4, 22, 5, 23, 12]$ and its end-effector follows the \textcolor{cyan}{cyan} path $\Teff$. 
    	An animation is available at \href{https://cobra.cps.cit.tum.de/ICRA24/example}{cobra.cps.cit.tum.de/ICRA24/example}.}
    \label{fig:solvingTask}
\end{figure}

Within the motion planning community, only a few benchmarks are established, e.g., by the creators of the Open Motion Planning Library (OMPL) \cite{Cohen2012, Moll2015}\footnote{\href{https://web.archive.org/web/20230705124355/https://plannerarena.org/}{plannerarena.org/}, accessed on Feb. 29th, 2024}, or Parasol\footnote{\href{https://web.archive.org/web/20170503025420/https://parasol.tamu.edu/groups/amatogroup/benchmarks/}{parasol.tamu.edu/groups/amatogroup/benchmarks/}, accessed on Feb. 29th, 2024}.
These are either limited to simple point-to-point planning or only contain abstract planning problems without a specific application.
In contrast, a benchmark suite specialized in a specific use case is CommonRoad for autonomous driving \cite{Althoff2017} or MotionBenchMaker for manipulator motion planning \cite{Chamzas2022}.
However, no benchmark suite exists for evaluating optimal robots or modular robot assemblies for a given task. 
We provide the first benchmark suite to compare robots and modular robot assemblies in different real-world environments for various cost functions.
An example solution to a benchmark task defined in a 3D-scanned environment is shown in Fig.~\ref{fig:solvingTask}.

In our literature overview, we survey robot descriptions, common robotic tasks in industry, and modular robot configuration optimization; we also state the objectives of our benchmark suite.

\subsection{Related Work}

\subsubsection{Robot (Task) Description}

An extensive and continuously updated overview of domain-specific languages for robotics is provided in \cite{Nordmann2016}. However, we have not found a common framework to describe (modular) robots, tasks, and cost functions.
Nonetheless, \cite{Bordignon2011} inspired the extension of our previous module description \cite{Althoff2019} detailed in Sec.~\ref{ssec:robotModel}.

MoveIt! \cite{Chitta2012} is a common software stack for robotic motion planning, which integrates OMPL \cite{Sucan2012} for planning within the Robot Operating System (ROS) \cite{ros}. 
On top of MoveIt, the work in \cite{Gorner2019} creates solutions to many robotic tasks with interdependent sub-tasks. 
Due to the deep integration in MoveIt, their task description  is not portable.

\subsubsection{Typical Robotic Tasks}
\label{ssec:RobotTasks}

Our \textbf{Co}mposable \textbf{B}enchmark (suite) for \textbf{R}obotics \textbf{A}pplications (CoBRA) intends to capture the variety of real-world applications a robot might encounter. 
Based on an analysis of market shares of different applications in robotics from 2023 \cite{InternationalFederationofRoboticsIFR2023}, these are still mainly in manufacturing (see Tab.~\ref{tab:robTasksInIndustry}).
In \cite{Wilson2014}, \cite[Ch. 54.4]{Siciliano2016} we found descriptions of the type of actions mainly required for each of these tasks. 
Most tasks can be broken down into point-to-point movements (PTP), (fixed) Cartesian trajectories, or force control. 
Some of these processes may need additional feedback, e.g., from vision, which is needed for grasp planning, quality assessment, and reworking.

\begin{table}[]
    \vspace*{1.7mm}  
    \caption{Common robot tasks in industry with their market share and predominant actions \cite[p.~14]{InternationalFederationofRoboticsIFR2023}.}
    \centering
    \begin{tabular}{@{}lSl@{}}
        \toprule
        Task & {Market Share} & Predominant Actions \\
        \midrule
        Handling / Tending & $\SI{48.1}{\percent}$ & PTP, Trajectory \\
        Welding & $\SI{15.7}{\percent}$ & PTP, Trajectory \\
        Assembly & $\SI{11.0}{\percent}$ & PTP, Trajectory, Force Control \\
        Dispensing & $\SI{5.1}{\percent}$ & Trajectory \\
        Predominant Processing & $\SI{1.1}{\percent}$ & Trajectory, Force Control \\
        Other \& Unspecified & $\SI{19.0}{\percent}$ & (Indeterminable) \\
        \bottomrule
    \end{tabular}
    \label{tab:robTasksInIndustry}  
\end{table}

\subsubsection{Optimizing Modular Robots}

One key aspect we address with our benchmark suite is modular and/or reconfigurable robotics \cite[Ch. 22.2]{Siciliano2016}, where even small module sets can be used to assemble millions of possible robots\cite{Liu2020}. 
Recent solutions to find optimal assemblies have combined hierarchical elimination with kinematic restrictions \cite{Liu2020, Althoff2019}, genetic algorithms \cite{Icer2017}, heuristic search \cite{Ha2018}, and reinforcement learning \cite{Whitman2020}.
The above-mentioned methods use modules and create modular robots with significantly different properties. 
Comparisons of the above-mentioned optimization methods are either not performed at all (e.g., \cite{Althoff2019, Ha2018}) or use re-implementations of other work and thereby impair comparability \cite{Whitman2020}. 
Analyzing these optimizers on a shared set of tasks and robots enables fair comparisons.

\subsection{Contributions}

We introduce CoBRA, a composable benchmark (suite) for robotics applications, which tests the selection, synthesis, and programming of robots based on well-defined environments, (modular) robot models, and task descriptions.
Perception of and reaction to an unexpected environment are intentionally left out for now to focus on evaluating robot assemblies for the intended industrial tasks with known and controlled environments.

Our benchmark suite strives for the same properties as \cite{Althoff2017}:

\begin{itemize}
 \item\textbf{Unambiguous}: The entire benchmark settings can be referred to by a unique identifier, and all details are provided in manuals on our website.
 \item\textbf{Composable}: Splitting each benchmark into components (robot model, tasks, and cost functions) allows one to easily compose new benchmarks by recombining existing components.
 \item\textbf{Representative}: Our benchmark suite contains real 3D environments and hand-crafted tasks covering various robotic tasks identified in the literature.
 \item\textbf{Portable}: All components of our benchmark suite are described in the JSON format, which makes it independent of specific platforms and programming languages. An interface to URDF eases the integration of the robots into different workflows.
 \item\textbf{Scalable}: Tasks can describe simple pick-and-place tasks as well as complex tasks, such as polishing a surface. Our robot description is valid for standard serial kinematics and extendable to modular robots with (multiple) closed kinematic chains and complex modules with multiple bodies, joints, connection points, bases, and end effectors.
 \item\textbf{Open}: Our benchmark suite is published in an open format on our website free of charge. Benchmark suggestions from the community are welcome.
 \item\textbf{Independent}: The benchmark descriptions are independent of any tools or products, making them suitable for any robot design and task.
\end{itemize}

The remaining paper is organized as follows: In Sec.~\ref{sec:problemStatement}, we provide the formal definitions of the optimization problem to be solved in each benchmark.
Next, in Sec.~\ref{sec:robots}, we describe the implementation of the robots, cost functions, and task descriptions. 
Lastly, in Sec.~\ref{sec:example}, we state our current approach for task generation and provide an example.

\section{Task Description and Problem Statement}
\label{sec:problemStatement}

Every benchmark $\benchmark$ comprises a set of robot modules $\modules$ (possibly a single robot), a cost function with ID $\costid$, and a task $\task$. 
Each task itself includes obstacles~$\workspace_{occ}$, a set of constraints~$\constraints$, and a set of goals~$\G$.

Within a given release of the benchmark suite, a benchmark $\benchmark$ can be written as 

\begin{equation}
    \benchmark = \modules:\costid:\task.
    \label{equ:singleBenchmark}
\end{equation}

An example is $\modules$ = Panda, $\costid$ = cyc (cycle time), $\task$ = Factory1 resulting in the benchmark ID \texttt{Panda:cyc:Factory1}. 
As in CommonRoad \cite{Althoff2017}, one can integrate individual or novel module sets, cost functions, or tasks and denote them by the keyword \textit{IND} or indicate modification by prefixing them with \textit{M-}.
Both \textit{IND} and \textit{M-} need auxiliary explanations in the accompanying publications.
Collaborative robot tasks can also be described, where $n \in \naturals^+$ robots work in a common task $\task$ and each robot fulfills tasks with its individual cost function $\costid_1$ to $\costid_n$:

\begin{equation}
    \benchmark = [\modules_1, ..., \modules_n]:[\costid_1, ...,  \costid_n]:C\text{-}\task.
    \label{equ:multiRobot}
\end{equation}

Note that the prefix \textit{C-} of task $\task$ indicates that it is collaborative. The prefix \textit{M-} precedes \textit{C-}.

\subsection{Poses}
\label{ssec:Poses}

Our benchmarks use poses $\T \in \SEThree$\textsuperscript{\ref{fn:SE3}}.
We denote the rotation matrix $\rotM$ in $\T$ as $\rotM(\T) \in \SOThree$\footnote{$\SOThree$ is the special orthogonal group represented as $3 \times 3$ rotation matrices; $\SEThree = \reals^3 \times \SOThree$ represents both translations and rotations.\label{fn:SE3}} and the translation $\Transv$ in $\T$ as $\Transv(\T) \in \reals^3$.
With these, any point represented as $\Pointv_a \in \reals^3$ in frame $a$ is represented with respect to frame $b$ by: 
\begin{equation}
    \Pointv_b = \T_b^a \Pointv_a = \rotM(\T_b^a) \Pointv_a + \Transv(\T_b^a) \; \text{\footnotemark} .
\end{equation}
\footnotetext{We use points $\Pointv = [p_x, p_y, p_z]^T$ and their homogeneous extension $\Pointv = [p_x, p_y, p_z, 1]^T$ interchangeably so that $\Pointv_b = \T_b^a \Pointv_a = \begin{bmatrix}
\rotM_b^a & \Transv_b^a \\
0     & 1
\end{bmatrix} 
\begin{bmatrix}
   \Pointv_a \\ 1 
\end{bmatrix} $.} 
To constrain and judge the distance between poses, we use a notation similar to \cite[Sec.~IV.A]{Berenson2009a}:
For subsequent formalizations, we introduce the mapping $\Proj \colon \SEThree \mapsto \reals^n, n \in \naturals^+$ and the shorthand $\Diff{\T}{\Tdes} = \Proj(\Tdes^{-1} \T)$ to denote the difference between a pose $\T$ and a desired pose $\Tdes \in \SEThree$ after applying $\Proj$.
$\Proj$ can be any combination of the projections listed in Tab.~\ref{tab:poseProjection}, e.g., the default mapping $\Proj(\T) = [x(\T), y(\T), z(\T), \theta_R(\T)]^T$ contains the three Cartesian coordinates and the rotation angle about an axis for any pose~$\T$.

\begin{table}[]
    \vspace*{1.7mm}  
    \caption{Available projections to coordinates from \cite[Tab.~1.2]{Siciliano2016}.}
    \centering
    \begin{tabular}{@{}lll@{}}
        \toprule
         & Projection Name & Coordinates \\
        \midrule
        Translation $\Transv(\T)$ & Cartesian & $[x, y, z]$ \\
         & Cylindrical & $[r_{cyl}, \theta, z]$ \\
         & Spherical & $[r_{sph}, \theta, \phi]$ \\
        Rotation $\rotM(\T)$ & XYZ fixed & $[r, p, y]$ \\
         & ZYX Euler & $[\alpha, \beta, \gamma]$ \\
         & Axis-angle & $[n_x, n_y, n_z, \theta_R]$ \\
         & Quaternion & $[a, b, c, d]$ \\
        \bottomrule
    \end{tabular}
    \label{tab:poseProjection}
\end{table}

Most real-world tasks accept some tolerance in execution so that each coordinate $i$ can vary between a minimum $\Tolbound^{\max}_i \in \reals$ and maximum value $\Tolbound^{\max}_i \in \reals$, respectively, which we denote by the interval vector ${\Tolboundv}(\Tdes) = [\Tolbound^{\min}, \Tolbound^{\max}] \in \reals^{2 \times N}$.
We say $\T$ fulfills $\Tdes$ if $\Diff{\T}{\Tdes} \in \Tolboundv(\Tdes)$.
As an example, one may want to constrain the end effector to be upright (z-axis pointing upwards), which can be done by using a desired pose $\Tdes = \Ident_{4 \times 4}$ (four-by-four identity matrix) in the world frame, the projections $\Proj(\T) = [r(\T), p(\T)]^T$, and setting both intervals from $\Tolbound^{\min} = -10^{-3}$ to $\Tolbound^{\max} = 10^{-3}$.

\subsection{Hybrid Motion Planning Problem}
\label{ssec:hybMotPlan}

Our benchmark suite extends classical motion planning for robotics \cite{Lynch2017}, which only searches for a motion $\Jmotionv(\timenow)$ over time $\timenow$ satisfying a set of goals $\G$ and constraints $\constraints$, detailed later on. 
We extend this planning problem by 
\begin{itemize}
\item synthesizing robots from modules, resulting in a tuple of $n$ modules $\assembly = [\module{1}, ..., \module{n}],\, \module{i} \in \modules$ to be assembled for serial kinematics;\footnote{Extensions to parallel kinematics are discussed in \href{https://cobra.cps.cit.tum.de/ICRA24/robot}{cobra.cps.cit.tum.de \textrightarrow Documentation \textrightarrow Robot}.}
\item optimizing the base pose $\B \in \SEThree$ (see Sec.~\ref{ssec:Poses}).
\end{itemize}

For a (rigid) robot with $n \in \naturals^+$ joints at time $\timenow$, the state of the robot is given by the joint configuration $\Jposv(\timenow) \in \reals^n$ and its derivative $\Jvelv(\timenow)$.
The state changes by the commanded accelerations $\Jaccv(\timenow)$.
We combine both in the state and input vector $\Jmotionv(\timenow) = (\Jposv(\timenow), \Jvelv(\timenow), \Jaccv(\timenow))$.
If unnecessary, we omit the dependency on time.
Based on the information provided in the corresponding benchmark, we can construct a robot model, including its kinematics and dynamics \cite{Althoff2019}. 
This model includes functions to calculate the

\begin{itemize}
    \item end effector pose $\Teff(\Jposv, \assembly) \in \SEThree$ relative to the base $\B$ \cite{Althoff2019}; for inverse solutions we use \cite[Sec.~2.7]{Siciliano2016};
    \item robot occupancy $\occupancy(\Jposv, \B, \assembly) \subset \powerset(\reals^3)$, where $\powerset(\bullet)$ returns the power set of $\bullet$; 
    \item forward dynamics given control forces $\Jtorqv$ and the external wrench $\Fext(t)$: $\Jaccv = \dynamics(\Jposv, \Jvelv, \Jtorqv, \Fext, \B, M)$ \cite{Althoff2019};
    \item inv. dynamics $\Jtorqv = \dynamics^{-1}(\Jmotionv, \Fext, \B, \assembly)$ \cite[Ch.~3.5]{Siciliano2016}.
\end{itemize}

The functions defined above allow us to define cost functions $\costfun{\costid}$ for each cost ID $\costid$.
These evaluate a proposed solution with its base pose $\B$, module order $\assembly$, and a motion specified for the time interval $[\timezero, \timeend]$ (without loss of generality, we set $t_0 = \timezero$). 
For $\costfun{\costid}$ we add terminal costs $\Phi_\costid$ and the integral of running costs $L_\costid$:

\begin{multline}
    \costfun{\costid}(\Jmotionv(\timenow), \timeend, \B, \assembly) = 
    \Phi_\costid(\Jmotionv(\timezero), \Jmotionv(\timeend), \timeend, \B, \assembly) + \\
    \int_{\timezero}^{\timeend} L_\costid(\Jmotionv(\timeany), \timeany, \B, \assembly) \mathrm{d}\timeany.
\end{multline}

Formally, we define the \textit{hybrid motion planning problem} as finding a module order $\assembly^*$, base placement $\B^*$, and motion vector $\Jmotionv^*$ minimizing a given cost function $\costfun{\costid}$:
\begin{equation}
    [\assembly^*, \B^*, \Jmotionv^*] = 
    \argmin_{\assembly, \B, \Jmotionv} \costfun{\costid}(\Jmotionv(t), \timeend, \B, \assembly)
    \label{equ:hybOptim}
\end{equation}
subject to $\forall \timenow \in [\timezero, \timeend]$:
\begin{align} 
    \Jaccv &= \dynamics(\Jposv, \Jvelv, \Jtorqv, \Fext, \B, \assembly) \\
    \forall \constraint \in \constraints \colon \constraint (\Jmotionv, \timenow, \B, \assembly) &\leq 0.
\end{align}
Initially, the robot must satisfy all constraints in $\constraints$ (see Sec.~\ref{ssec:Constraint}) and be stationary, i.e., $\Jmotionv(\timezero) = [\Jposv(\timezero), \Zerov_n, \Zerov_n]$, where $\Zerov_n$ is a vector of $n$ zeros. 
Note that if the available module set $\modules$ only includes one valid robot, and $\B$ is restricted to a single pose, this hybrid motion planning problem is reduced to standard motion planning for robotic manipulators.
Subsequently, we will introduce our definitions for constraints in $\constraints$ and goals in $\G$.

\subsection{Constraints}
\label{ssec:Constraint}

We consider constraints that can be written as \mbox{$\constraint(\Jmotionv(\timenow), \timenow, \B, \assembly) \leq 0$}, noting that equality can be ensured by adding the partial constraints $\constraint[i] \leq 0 \wedge -\constraint[i] \leq 0$. 
To incorporate Boolean functions $b$ in our constraint formulation, we use the operator $\mathbb{I}(b)$, which evaluates to $-1$ if $b$ is true and otherwise to $1$. 
We split $\constraints$ into constraints holding for an entire task $\task$, named $\constraints_S$, and those applying to a single goal $\goal[i] \in \G$ named $\constraints_i$.
Additionally, we introduce the
\begin{itemize}
\item robot Jacobian $J(\Jposv) = \frac{\mathrm{d} \Teff(\Jposv)}{\mathrm{d} \Jposv}$;
\item velocities of the end effector $[\vtrans, \vrot] = J(\Jposv) \Jvelv$;
\item occupancy of link $\link$ from set of robot links $\links$ as $\tilde{\occupancy}(\link)$;
\item desired base pose with tolerance $\B_{\given}$.
\end{itemize}
For vectors $\vtrans \in \reals^n$, we denote the componentwise absolute value as $|\vtrans|$ and the 2-norm as $\|\vtrans\|_2$. 
The time to finish a single goal $\goal[i]$ is $\timefinal{i}$.
With these, we list the supported constraints to formalize $\constraints_S$ and $\constraints_i$ in Tab.~\ref{Tab:ConstrFunctions}.

\begin{table}
    \vspace*{1.7mm}  
    \caption{Available constraint functions.}
    \label{Tab:ConstrFunctions}
\centering
\begin{tabular}{p{28mm}p{48mm}}
        \toprule
         Constraint & Constraint Function \\
        \midrule
Joint position limits                           & $\Jposv \geq \Jposv_{\min} \wedge \Jposv \leq \Jposv_{\max}$                              \\[1ex]
Joint velocity and \newline acceleration limits & $|\Jvelv| \leq \Jvelv_{\max}, |\Jaccv| \leq \Jaccv_{\max}$      \\
Joint torque limits                    & $|\dynamics^{-1}(\Jmotionv, \Fext, \B, \assembly)| \leq \Jtorqv_{\max}$                   \\
        \midrule
End effector velocity                  & $\|\vtrans\|_2 \in [\velocity_{\min}, \velocity_{\max}],$ \newline $\|\vrot\|_2 \in [\omega_{\min}, \omega_{\max}]$ \\[4ex]
Keep $\Teff$ close to $\Tdes$ & $\Diff{\Teff}{\Tdes} \in \Tolboundv(\Tdes)$ \\
        \midrule
Fulfill all goals & $\forall \goal[i] \in \G \, \exists \timefinal{i} \in [\timezero, \timeend]$ \\[1ex]
Fulfill all goals in order & $\forall \goal[i], \goal[j] \in \G,  i < j \colon \timefinal{i} < \timefinal{j}$ \\
        \midrule
No self-collision & $\forall \link, \link' \in \links, \, \link \neq \link' \colon \tilde{\occupancy}(\link) \cap \tilde{\occupancy}(\link') = \emptyset$ \\[1ex]
No collision & $\occupancy(\Jposv, \B, \assembly) \cap \workspace_{occ} = \emptyset$ \\
        \midrule
Valid base pose $\B_{\given}$ & $\Diff{\B}{\B_{\given}} \in \Tolboundv(\B_{\given})$ \\[1ex]
Valid assembly & see Sec.~\ref{ssec:robotModel} \\
         \bottomrule
\end{tabular}
\end{table}

\subsection{Goals}
\label{ssec:Goals}

In every task $\task$, all robots have to fulfill a set of $n$ goals $\G = \{\goal[1], ..., \goal[n]\}$.
As some goals are not fulfilled instantaneous, we introduce the duration operator $\timeduration(\bullet)$ which returns the length of $\bullet$, such as a goal or trajectory, in seconds.
Each goal $\goal[i] \in \G$ contains a set of constraint functions $\constraints_i$, a termination condition $g_i(\Jmotionv, \timenow, \B, \assembly)$, and an execution duration $\timeduration[{\goal[i]}]$.
For convenience, we order the goals such that $\goal[i]$ is fulfilled after $\goal[i-1]$.
A goal $\goal[i]$ is fulfilled at time $\timefinal{i}$ if $g_i(\Jmotionv, \timenow, \B, \assembly)$ evaluates to true and its constraints have been satisfied within its duration $\timeduration[{\goal[i]}]$: ${\forall \timeany \in [\timefinal{i}-\timeduration[{\goal[i]}], \timefinal{i}]}, {\forall \constraint \in \constraints_i}\colon {\constraint(\Jmotionv(\timeany), \timeany, \B, \assembly) \leq 0}$.

We introduce a small deviation $\epsilon = 10^{-3}$ and a Cartesian trajectory $\T(\timenow)$ describing desired poses for each time step within $[\timezero, \timeduration[{\T(\timenow)]}]$.
Some goals may have an explicit duration, such as a pause for $\timeduration$ seconds.
Additionally, $\workspace_A \subset \powerset(\reals^3)$ is a space to temporarily leave, e.g., the workspace of a machine (see Sec.~\ref{sec:tasks}). 
With these prerequisites, we expect that termination conditions $g$ for most tasks given in Tab.~\ref{tab:robTasksInIndustry} can be composed of the predicates in Tab.~\ref{Tab:GoalFunctions}. 

\begin{table}
    \vspace*{1.7mm}  
    \caption{Available goal predicates.}
    \label{Tab:GoalFunctions}
\centering
\begin{tabular}{p{22mm}p{56mm}}
        \toprule
         Goal & Goal Predicate \\
        \midrule
$\at(\Tdes, \Jposv)$ & $\Diff{\B \, \Teff(\Jposv,  \assembly)}{\Tdes} \in \Gamma(\Tdes)$ \\[1ex]
$\reach(\Tdes, \Jposv, \Jvelv, \Jaccv)$ & $\at(\Tdes, \Jposv) \wedge \|\Jvelv\|_2 \leq \epsilon \wedge \|\Jaccv\|_2 \leq \epsilon$      \\[1ex]
$\followed(\T(\timenow), \Jposv, \timenow) $ & 
$\forall \timeany \in [\timezero, \timeduration[{\T(\timenow)}]] \colon \newline
\at(\T(\timeduration[{\T(\timenow)}] - \timeany), \Jposv(\timenow-\timeany))$ \\
		\midrule
$\returnto(\goal[i], \timenow)$ & $\|\Jposv(\timenow) - \Jposv(\timefinal{i})\|_2 \leq \epsilon \wedge \|\Jvelv(\timenow)-\Jvelv(\timefinal{i})\|_2 \leq \epsilon \wedge$ \newline $\|\Jaccv(\timenow)-\Jaccv(\timefinal{i})\|_2 \leq \epsilon$ \\[4ex]
$\pause(\Jposv, \timeduration, \timenow)$ & $\forall \timeany \in [\timenow-\timeduration, \timenow]\colon \|\Jposv(\timeany) - \Jposv(\timenow)\|_2 \leq \epsilon$ \\[1ex]
$\leftw(\workspace_A, \Jposv, \timeduration, \timenow)$ & $\forall \timeany \in [\timenow-\timeduration, \timenow]\colon \occupancy(\Jposv(\timeany), \B, \assembly) \cap \workspace_A = \emptyset$ \\
         \bottomrule
\end{tabular}
\end{table}

With respect to Tab.~\ref{tab:robTasksInIndustry}, \textit{at} and \textit{reach} model simple PTP tasks.
The predicate \textit{followed} allows one to define more complex trajectories to be followed with precise timing.
\textit{ReturnTo}, \textit{pause}, and \textit{left} specify synchronization with external events, such as a CNC mill processing a placed part, without restricting the robot to specific positions.

\section{Implementation}
\label{sec:robots}

The next three subsections provide implementation details for modeling robots, define cost functions, and describe how tasks and proposed solutions are stored.

\subsection{Robots}
\label{ssec:robotModel}

We propose a model description similar to the universal robot description format (URDF)\footnote{\href{https://wiki.ros.org/urdf/XML}{wiki.ros.org/urdf/XML}, accessed on Feb. 29th, 2024} and alter it such that \textit{modules} $\module{i}$, rather than assembled robots, are the basic entities.
Modules generalize our previous description from \cite{Althoff2019}, considering more than two rigid bodies $\body{1, 2}$ and a single joint $\joint{1}$.
They also integrate \textit{connectors} $\connector$, as introduced by \cite{Bordignon2011}, to explicitly model valid assemblies.

Each module $\module{i}$ has a unique \texttt{ID}$(\module{i})$ within a module set $\modules$ with a unique \texttt{name}$(\modules)$.
Any robot with module order $\assembly$ can, therefore, be described as a string \texttt{name$(\modules)$.[ID($\module{1}, \ldots, \module{N}$)]}, see Sec.~\ref{sec:example}.
CoBRA provides a web API to generate URDFs for any included robot\footnote{For example \href{https://cobra.cps.cit.tum.de/ICRA24/urdf}{cobra.cps.cit.tum.de/ICRA24/urdf} provides the URDF for the robot in Fig.~\ref{fig:solvingTask}.}.

\begin{figure}
    \centering
    \resizebox{0.45\textwidth}{!}{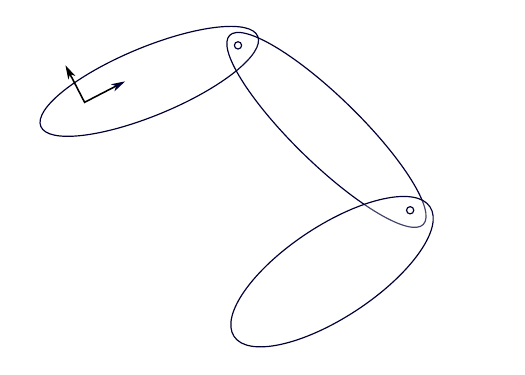}
    \caption{
        Sketch of a module consisting of three ellipsoidal bodies $\body{1}, \body{2}, \body{3}$, each with a body-fixed frame. 
        The bodies are connected via two joints $\joint{1}, \joint{2}$ (empty circles), and the joint transformation $\T_{\parentBody} \T_{\joint{1}}(\Jposv) \T_{\childBody}$ is shown between $\body{1}$ and $\body{2}$.
        Each body has a connector $\connector_1, \connector_2, \connector_3$, and the pose $\T^{\connector_2}_{\body{2}}$ for connector $\connector_2$ is shown. 
        Body $\body{3}$ displays the transformation $\T_{\body{3}}^{\com}$ to its center of mass and $\T_{\body{3}}^{\bodyOcc{3}}$ to the ellipse describing its occupancy $\bodyOcc{3}$.}
    \label{fig:module}
\end{figure}
\begin{figure}
    \vspace*{1.7mm}  
    \centering
    \resizebox{\columnwidth}{!}{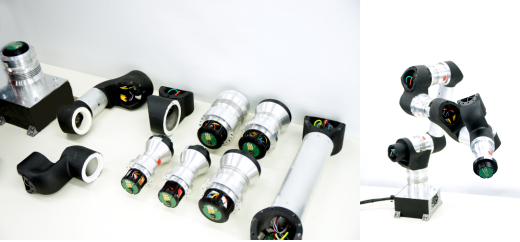}
    \caption{Modules (left) and links in the assembled robot (right) from the module set \texttt{PROMODULAR} with \textcolor{red}{red} lines marking the separation of bodies in a module and \textcolor{blue}{blue} lines separating a link in the assembled robot. Based on \cite[Fig.~1]{Liu2020}.}
    \label{fig:assembledModules}
\end{figure}

Similar to \cite{Bordignon2011, Althoff2019}, we compose modules of \textit{bodies} $\body{i}$ and \textit{joints}.
Fig.~\ref{fig:module} illustrates a single module, and Fig.~\ref{fig:assembledModules} is an example of real modules assembled into a robot.
Each body $\body{i}$ specifies the dynamics, e.g., the center of mass \textit{CoM}, and the occupancy $\bodyOcc{i}$, e.g., each ellipse in Fig.~\ref{fig:module}, of a rigid body. 
Additionally, each module details how to connect it to others via \textit{connectors} $\connector$.
Each connector contains lists of supported \textit{sizes} $\in \powerset(\reals)$ and \textit{types} $\in \powerset(\texttt{string})$.
Multiple sizes realize module designs where adjacent sizes can fit together, but not all combinations are valid; similarly, multiple types enable the construction of modules that fit with multiple connection designs.
Note that bodies from multiple modules are connected rigidly and together form a link, as known, e.g., from URDF; the joints within any module separate links.
We denote the set of links in the assembled robot as $\links$.

A robot assembly is \textit{valid} if connectors match, i.e., they share at least a common entry in their respective \textit{size}  and \textit{type} lists, and their \textit{genders} are both hermaphroditic (gender-less) or opposing (male/female).
This matching extends the one based on gender alone, as introduced in \cite{Bordignon2011}.
Additionally, each connector $\connector$ defines its \textit{pose} relative to the body frame $\body{i}$ as $\T^\connector_{\body{i}}$, shown for $\body{2}$ in Fig.~\ref{fig:module}.
If two connectors $\connector_{A, B}$ with pose $\T^{\connector_A}_{A}, \T^{\connector_B}_{B}$, relative to body $A$ and $B$, are connected during assembly, their x-axes align, and their z-axes are anti-parallel.
For each valid assembly of modules, we can generate kinematic, dynamic, and collision models\textsuperscript{\ref{fn:robotSpec}}.
As in \cite{Bordignon2011}, an unlimited number of connectors on a module enables us to also model robots with branching or closed kinematic chains\footnote{\href{https://cobra.cps.cit.tum.de/ICRA24/robot}{cobra.cps.cit.tum.de \textrightarrow Documentation \textrightarrow Robot}\label{fn:robotSpec}}.

A joint $\joint{i}$ connects two bodies, referred to as \textit{parent} $\parentBody$ and \textit{child} $\childBody$. 
Each joint $\joint{i}$ needs three transformations: $\T_{\parentBody}$ from the parent $\parentBody$ to the joint frame, the joint transformation $\T_{\joint{i}}(\Jpos, \mathrm{type})$ and $\T_\childBody$, from the joint to the child $\childBody$.
The most common \textit{type} of joint is \textit{revolute}, where $\T_{\joint{i}}(\Jpos, \mathrm{revolute})$ is a single rotation about the z-axis of the joint frame by the angle $\Jpos$.
An example is visualized between body $\body{1}$ and $\body{2}$ in Fig.~\ref{fig:module}.
In addition, we also model the drivetrain dynamics for joints, with gear ratio $\gearRatio$, motor side inertia $\rotorInertia$, Coulomb and viscous friction $\fcoulomb, \fviscous$, respectively.
These result in an additional torque $\Jtorq_i$ in joint $\joint{i}$ \cite[Ch.~7]{Siciliano2009}
\begin{equation}
    \Jtorq_i = \rotorInertia \gearRatio^2 \, \Jacc_i + \fviscous \, \Jvel_i + \fcoulomb \, \sign(\Jvel_i).
\end{equation}

Currently, we provide the set of robot modules $\mathtt{IMPROV}$ based on Schunk's LWA 4P \cite{Althoff2019} and $\mathtt{PROMODULAR}$ from \cite{Liu2020}.
Additionally, we provide standard industrial robots, such as Franka Emika's Panda, with dynamic parameters identified in \cite{Gaz2019} as a point of reference.

\subsection{Cost Functions}
\label{sec:costs}

We include cost functions used in (modular) robot optimization from the literature in CoBRA. 
In general, they can be split into \textit{atomic} functions that map a robot and/or its motion to a scalar value or compound functions that weigh multiple atomic functions.
A detailed description of all available costs can be found on our website\footnote{\href{https://cobra.cps.cit.tum.de/ICRA24/solution}{cobra.cps.cit.tum.de \textrightarrow Documentation \textrightarrow Solution}\label{fn:solutionDoc}}.

In our example in Sec.~\ref{sec:example}, we use atomic costs describing the robot mass ($\massJ$) in kilograms, number of actuated joints ($\numJ$), the time it takes to move the robot ($\timeJ$) in seconds and the mechanical energy required ($\mechE$) in joules. 
In addition, we evaluated two combined cost functions:
\texttt{Liu2} weighs mechanical energy and execution time so that $\Liu = \mechE + \SI{0.2}{} \, \timeJ$ \cite[Sec.~IV~B]{Liu2020} and \texttt{Whit2} the number of actuated joints and the robot mass so that $\Whit = 0.025 \numJ + \SI{0.1}{} \, \massJ$ \cite[Tab.~1]{Whitman2020}.
Any other weighted sum of $N$ cost functions can be abbreviated as $[(\costfun{1} | w_1), ..., (\costfun{N} | w_N)] $ to define a total $\costfun{\costid} = \sum_{i=1}^N w_i \costfun{i}$.

\subsection{Tasks}
\label{sec:tasks}

\begin{figure}
    \vspace*{1.7mm}  
	\small
	\dirtree{%
	  .1 [1] task.
	    .2 [1] header.
	      .3 [1] taskID: string.
	      .3 [1] version: string.
	      .3 [1..] author: List[string].
	      .3 [0..] tags: string = [].
	      .3 [1] date: string yyyy-MM-dd.
	      .3 [0..1] gravity: $\reals^3$ = $ [0, 0, -9.81]^T $.
		.2 [0..] obstacles.
	    .2 [0..] constraints.
		.2 [1..] goals.
		  .3 [1] ID: string.
		  .3 [1] type: \{at, reach, returnTo, pause, follow, leave\}.
		  .3 [1] parameter (specific to \textit{type}).
	}
	\caption{Abbreviated structure of the task object. For each attribute, we provide ranges for the number of allowed elements. Fields with a lower bound of \texttt{0} elements are optional and come with default values. Fields that allow more than one element are arrays. For each primitive field, the type is given after a colon. The complete version, including all fields and drafted extensions, is published on our website\textsuperscript{\ref{fn:taskSpec}}.}
	\label{fig:TaskStructure}
\end{figure}

The structure of a task is shown in Fig.~\ref{fig:TaskStructure}, describing its obstacles $\workspace_{occ}$, constraints $\constraints$ (Sec.~\ref{ssec:Constraint}) and goals $\G$ (Sec.~\ref{ssec:Goals}). 
A task ID and a benchmark version uniquely identify each task.
Additionally, it contains contact information from the authors, tags for semantic searches, and a date of publishing. 
The complete description can be found on our website\footnote{\href{https://cobra.cps.cit.tum.de/ICRA24/task}{cobra.cps.cit.tum.de \textrightarrow Documentation \textrightarrow Task}\label{fn:taskSpec}}.

Obstacles are stationary parts of the environment, such as machines or columns, that are placed at a pose relative to the world frame and are represented by collision and optional visual geometries.
Those two geometries enable efficient collision checking on simpler over-approximations while keeping fidelity for visualization if needed.

For every constraint and goal, the task must reference the specific functions in Sec.~\ref{ssec:Constraint}, \ref{ssec:Goals} via a type and provide their parameters, such as
\begin{itemize}
\item poses (with tolerances) $\T$ (or an array of these to follow)\textsuperscript{\ref{fn:taskSpec}});
\item references $\goal[i]$ to other goals via their ID;
\item scalars, such as, duration $\timeduration$ for pause or $\velocity_{\min}, \velocity_{\max}$ for end-effector constraints;
\item arrays of goal IDs to fulfill in order;
\item regions $\workspace_A$, which can be specified as any geometry, as defined for an obstacle.
\end{itemize}

\begin{figure}
    \vspace*{1.7mm}  
	\small
	\dirtree{%
	  .1 [1] Solution. 
	    .2 [1] taskID: string.
	    .2 [1] version: string.
	    .2 [1] costFunction: string.
	    .2 [1] moduleSet: string.
	    .2 [1..] moduleOrder: module\_id.
	    .2 [1] basePose: pose.
		.2 [1] trajectory of state and input.
		  .3 [1] t: $\reals^N$.
		  .3 [1] q: $\reals^{N \times DoF}$.
		  .3 [1] dq: $\reals^{N \times DoF}$.
		  .3 [1] ddq: $\reals^{N \times DoF}$.
		  .3 [1] goal2time: Dict(goal\_ID $\to$ time).
	}
	\caption{Abbreviated structure of the solution object for a robot with a single kinematic chain following the notation from Fig.~\ref{fig:TaskStructure}. The complete version is published on our website\textsuperscript{\ref{fn:solutionDoc}}.}
	\label{fig:SolutionStructure}
\end{figure}

The solution structure for a robot with a single kinematic chain is provided in Fig.~\ref{fig:SolutionStructure}. 
It specifies the task $\task$ by its ID and benchmark version, the ID of the used cost function $\costid$, and a robot via module set $\modules$, module order $\assembly^*$, and base pose $\B^*$.
The solving state and input vector $\Jmotionv^*$ as defined in \eqref{equ:hybOptim} is also given. 
It contains a sequence of $N$ samples of $\Jmotionv = [\Jposv, \Jvelv, \Jaccv]$ at times in \texttt{t}. 
For each goal in a task $\goal[i] \in \G$, \texttt{goal2time} states its final time $\timefinal{i}$.

Any solution can be submitted to our website where we check whether it solves the specified task $\task$ while satisfying the constraint set $\constraints$. 
Valid solutions will be published together with the provided author information and can be queried, e.g., for used cost functions or final cost.

\section{Numerical Example}
\label{sec:example}

Following \eqref{equ:singleBenchmark}, we compare solutions to benchmarks with 

\begin{itemize}
    \item module sets $\modules \in \{ \mathtt{PROMODULAR}, \mathtt{Panda}, \mathtt{IMPROV} \}$,
    \item cost function IDs $\costid \in \{ \mathrm{mechE}, \mathrm{Whit2}, \mathrm{Liu2} \}$, and
    \item a single task $\task = \mathtt{Liu2020/Case2b/Schunk\_IWB\_}$ $\mathtt{CTX\_300\_linear\_0}$.
\end{itemize}

Within $\task$, the goals $\G$ require the robot to  move in and out of a 3D-scanned CNC machine. 
We include \textit{joint limits} (in position and torque), \textit{no (self-)collisions}, and a fixed \textit{goal order} as constraints $\constraints$ (Sec.~\ref{ssec:Constraint}).
An example solution is shown in Fig.~\ref{fig:solvingTask} and all solutions are on our website\footnote{\href{https://cobra.cps.cit.tum.de/ICRA24/example}{cobra.cps.cit.tum.de/ICRA24/example}}.

Tab.~\ref{tab:costCompar} summarizes the minimal costs $\costfun{i}$ of the generated solutions for a robot from each set $\modules$ over ten generated trajectories.
All solution trajectories were generated with OMPL's RRT-Connect implementation\footnote{\href{https://ompl.kavrakilab.org/classompl_1_1geometric_1_1RRTConnect.html}{ompl.kavrakilab.org/classompl\_1\_1geometric\_1\_1RRTConnect.html}, accessed on Feb. 29th, 2024} and adhere to the constraints given in $\task$.

\begin{table}[]
    \vspace*{1.7mm}  
    \centering
    \caption{Minimal solution costs for the example task with different robots.}
    \begin{tabular}{@{}llll@{}}
    \toprule
    Cost $\costfun{i}$ (from \ref{sec:costs}) & $\mathtt{PROMODULAR}$ & $\mathtt{Panda}$ & $\mathtt{IMPROV}$ \\
    \midrule
    $\mechE$ & $\SI{1194.6}{}$ & $\SI{255.7}{}$ & $\SI{383.5}{}$ \\
    $\Whit$ & $\SI{2.085}{}$ & $\SI{1.84}{}$ & $\SI{1.522}{}$ \\
    $\Liu$ & $\SI{1198.8}{}$ & $\SI{260.7}{}$ & $\SI{386.4}{}$ \\
    \bottomrule
    \end{tabular}
    \label{tab:costCompar}
\end{table}

\section{Conclusions} \label{sec:conclusions}

This paper addressed the problem of finding optimal robotic solutions for industrial tasks using conventional and modular robots.
We propose a novel framework that integrates motion planning and modular robot optimization, which together can be evaluated on a set of realistic tasks in our benchmark suite CoBRA available at \href{https://cobra.cps.cit.tum.de}{cobra.cps.cit.tum.de}. 
It is a place to share the generated solutions to the proposed tasks and distribute novel tasks highlighting the properties of different optimizers.
The documentation of CoBRA includes detailed descriptions of the abstract objects described within this paper.

Specifically, CoBRA focuses on industrial settings with well-known environments based on synthesized obstacles or 3D scans of actual factory environments.
In particular, we include the inherent flexibility in many tasks, such as rotational symmetries of tools, tolerances in execution, or flexibility in the position of the robot's base, into the motion planning problem.
Executable robot models are available using Timor \cite{Kuelz22}, including kinematics, dynamics, and collision checking.

\section*{Acknowledgements}
This work was supported by the ZIM project on energy- and wear-efficient trajectory generation (grant no. ZF4086011PO8) and the EU's Horizon 2020 project CONCERT (grant no. 101016007).
We also thank the GrabCAD community and \cite{Gaz2019} for their published assets and robot models.

\bibliography{IEEEabrv,mmayerReferences_clean}
\bibliographystyle{IEEEtranWithoutAddress}

\end{document}

%% file: figures/compareIn3DScanNew.pdf_tex
\begingroup%
  \makeatletter%
  \providecommand\color[2][]{%
    \errmessage{(Inkscape) Color is used for the text in Inkscape, but the package 'color.sty' is not loaded}%
    \renewcommand\color[2][]{}%
  }%
  \providecommand\transparent[1]{%
    \errmessage{(Inkscape) Transparency is used (non-zero) for the text in Inkscape, but the package 'transparent.sty' is not loaded}%
    \renewcommand\transparent[1]{}%
  }%
  \providecommand\rotatebox[2]{#2}%
  \newcommand*\fsize{\dimexpr\f@size pt\relax}%
  \newcommand*\lineheight[1]{\fontsize{\fsize}{#1\fsize}\selectfont}%
  \ifx\svgwidth\undefined%
    \setlength{\unitlength}{244.79714701bp}%
    \ifx\svgscale\undefined%
      \relax%
    \else%
      \setlength{\unitlength}{\unitlength * \real{\svgscale}}%
    \fi%
  \else%
    \setlength{\unitlength}{\svgwidth}%
  \fi%
  \global\let\svgwidth\undefined%
  \global\let\svgscale\undefined%
  \makeatother%
  \begin{picture}(1,1.25614215)%
    \lineheight{1}%
    \setlength\tabcolsep{0pt}%
    \put(0,0){\includegraphics[width=\unitlength,page=1]{compareIn3DScanNew.pdf}}%
    \put(0.55480961,0.91144156){\color[rgb]{0,0,0}\makebox(0,0)[lt]{\lineheight{1.25}\smash{\begin{tabular}[t]{l}\colorbox{white}{Goal 2}\end{tabular}}}}%
    \put(0.72434586,0.76166677){\color[rgb]{0,0,0}\makebox(0,0)[lt]{\lineheight{1.25}\smash{\begin{tabular}[t]{l}\colorbox{white}{Goal 4}\end{tabular}}}}%
    \put(0.41670766,0.41090405){\color[rgb]{0,0,0}\makebox(0,0)[lt]{\lineheight{1.25}\smash{\begin{tabular}[t]{l}\colorbox{white}{Goal 1}\end{tabular}}}}%
    \put(0.294334,0.14645021){\color[rgb]{0,0,0}\makebox(0,0)[lt]{\lineheight{1.25}\smash{\begin{tabular}[t]{l}\colorbox{white}{Goal 3}\end{tabular}}}}%
  \end{picture}%
\endgroup%

%% file: figures/moduleExample.pdf_tex
\begingroup%
  \makeatletter%
  \providecommand\color[2][]{%
    \errmessage{(Inkscape) Color is used for the text in Inkscape, but the package 'color.sty' is not loaded}%
    \renewcommand\color[2][]{}%
  }%
  \providecommand\transparent[1]{%
    \errmessage{(Inkscape) Transparency is used (non-zero) for the text in Inkscape, but the package 'transparent.sty' is not loaded}%
    \renewcommand\transparent[1]{}%
  }%
  \providecommand\rotatebox[2]{#2}%
  \newcommand*\fsize{\dimexpr\f@size pt\relax}%
  \newcommand*\lineheight[1]{\fontsize{\fsize}{#1\fsize}\selectfont}%
  \ifx\svgwidth\undefined%
    \setlength{\unitlength}{252bp}%
    \ifx\svgscale\undefined%
      \relax%
    \else%
      \setlength{\unitlength}{\unitlength * \real{\svgscale}}%
    \fi%
  \else%
    \setlength{\unitlength}{\svgwidth}%
  \fi%
  \global\let\svgwidth\undefined%
  \global\let\svgscale\undefined%
  \makeatother%
  \begin{picture}(1,0.71428571)%
    \lineheight{1}%
    \setlength\tabcolsep{0pt}%
    \put(0,0){\includegraphics[width=\unitlength,page=1]{moduleExample.pdf}}%
    \put(0.1393317,0.48384252){\color[rgb]{0,0,0}\makebox(0,0)[lt]{\lineheight{1.25}\smash{\begin{tabular}[t]{l}$\body{1}$\end{tabular}}}}%
    \put(0,0){\includegraphics[width=\unitlength,page=2]{moduleExample.pdf}}%
    \put(0.2523227,0.44904441){\color[rgb]{0,0,0}\makebox(0,0)[lt]{\lineheight{0.20000094}\smash{\begin{tabular}[t]{l}\begin{tabular}[t]{c}Male\\connector $\connector_{1}$\end{tabular}\end{tabular}}}}%
    \put(0,0){\includegraphics[width=\unitlength,page=3]{moduleExample.pdf}}%
    \put(0.5632104,0.45370633){\color[rgb]{0,0,0}\makebox(0,0)[lt]{\lineheight{1.25}\smash{\begin{tabular}[t]{l}$\body{2}$\end{tabular}}}}%
    \put(0.63536418,0.26204855){\color[rgb]{0,0,0}\makebox(0,0)[lt]{\lineheight{1.25}\smash{\begin{tabular}[t]{l}$\body{3}$\end{tabular}}}}%
    \put(0,0){\includegraphics[width=\unitlength,page=4]{moduleExample.pdf}}%
    \put(0.48167492,0.65363251){\color[rgb]{0,0,0}\makebox(0,0)[lt]{\lineheight{1.25}\smash{\begin{tabular}[t]{l}$\T_{\joint{1}}(q)$\end{tabular}}}}%
    \put(0.82537545,0.32261359){\color[rgb]{0,0,0}\makebox(0,0)[lt]{\lineheight{1.25}\smash{\begin{tabular}[t]{l}Joint $\joint{2}$\end{tabular}}}}%
    \put(0,0){\includegraphics[width=\unitlength,page=5]{moduleExample.pdf}}%
    \put(0.26870803,0.57336046){\color[rgb]{0,0,0}\makebox(0,0)[lt]{\lineheight{1.25}\smash{\begin{tabular}[t]{l}$\T_\parentBody$\end{tabular}}}}%
    \put(0.66747556,0.42632066){\color[rgb]{0,0,0}\makebox(0,0)[lt]{\lineheight{1.25}\smash{\begin{tabular}[t]{l}$\T_{\body{2}}^{\connector_2}$\end{tabular}}}}%
    \put(0,0){\includegraphics[width=\unitlength,page=6]{moduleExample.pdf}}%
    \put(0.50320131,0.56961991){\color[rgb]{0,0,0}\makebox(0,0)[lt]{\lineheight{1.25}\smash{\begin{tabular}[t]{l}$\T_\childBody$\end{tabular}}}}%
    \put(0,0){\includegraphics[width=\unitlength,page=7]{moduleExample.pdf}}%
    \put(0.63890884,0.15191271){\color[rgb]{0,0,0}\makebox(0,0)[lt]{\lineheight{1.25}\smash{\begin{tabular}[t]{l}$\T_{\body{3}}^{\bodyOcc{3}}$\end{tabular}}}}%
    \put(0.54408362,0.11953352){\color[rgb]{0,0,0}\makebox(0,0)[lt]{\lineheight{1.25}\smash{\begin{tabular}[t]{l}$\T_{\body{3}}^{\com}$\end{tabular}}}}%
    \put(0.7370056,0.49770759){\color[rgb]{0,0,0}\makebox(0,0)[lt]{\lineheight{0.20000094}\smash{\begin{tabular}[t]{l}Female \\connector $\connector_2$\end{tabular}}}}%
    \put(0.52176429,0.2924108){\color[rgb]{0,0,0}\makebox(0,0)[lt]{\lineheight{0.20000094}\smash{\begin{tabular}[t]{l}$\connector_{3}$\end{tabular}}}}%
    \put(0,0){\includegraphics[width=\unitlength,page=8]{moduleExample.pdf}}%
    \put(0.20932413,0.19575023){\makebox(0,0)[lt]{\lineheight{1.25}\smash{\begin{tabular}[t]{l}Occupancy $\bodyOcc{3}$\end{tabular}}}}%
    \put(0.07179113,0.42428038){\makebox(0,0)[lt]{\lineheight{1.25}\smash{\begin{tabular}[t]{l}$\bodyOcc{1}$\end{tabular}}}}%
    \put(0.51479112,0.39385531){\makebox(0,0)[lt]{\lineheight{1.25}\smash{\begin{tabular}[t]{l}$\bodyOcc{2}$\end{tabular}}}}%
  \end{picture}%
\endgroup%

%% file: figures/robotNomenclature.pdf_tex
\begingroup%
  \makeatletter%
  \providecommand\color[2][]{%
    \errmessage{(Inkscape) Color is used for the text in Inkscape, but the package 'color.sty' is not loaded}%
    \renewcommand\color[2][]{}%
  }%
  \providecommand\transparent[1]{%
    \errmessage{(Inkscape) Transparency is used (non-zero) for the text in Inkscape, but the package 'transparent.sty' is not loaded}%
    \renewcommand\transparent[1]{}%
  }%
  \providecommand\rotatebox[2]{#2}%
  \newcommand*\fsize{\dimexpr\f@size pt\relax}%
  \newcommand*\lineheight[1]{\fontsize{\fsize}{#1\fsize}\selectfont}%
  \ifx\svgwidth\undefined%
    \setlength{\unitlength}{249.59998904bp}%
    \ifx\svgscale\undefined%
      \relax%
    \else%
      \setlength{\unitlength}{\unitlength * \real{\svgscale}}%
    \fi%
  \else%
    \setlength{\unitlength}{\svgwidth}%
  \fi%
  \global\let\svgwidth\undefined%
  \global\let\svgscale\undefined%
  \makeatother%
  \begin{picture}(1,0.46534665)%
    \lineheight{1}%
    \setlength\tabcolsep{0pt}%
    \put(0,0){\includegraphics[width=\unitlength,page=1]{robotNomenclature.pdf}}%
    \put(0.41858914,0.42069673){\color[rgb]{0,0,0}\makebox(0,0)[lt]{\begin{minipage}{0.53335338\unitlength}\raggedright Motor module\\ with two bodies\end{minipage}}}%
    \put(0.71793446,0.41270337){\color[rgb]{0,0,0}\makebox(0,0)[lt]{\begin{minipage}{0.53335338\unitlength}\raggedright Robot link\end{minipage}}}%
    \put(0,0){\includegraphics[width=\unitlength,page=2]{robotNomenclature.pdf}}%
    \put(0.00313395,0.0930296){\color[rgb]{0,0,0}\makebox(0,0)[lt]{\begin{minipage}{0.53335338\unitlength}\raggedright Module with\\ single body\end{minipage}}}%
    \put(0,0){\includegraphics[width=\unitlength,page=3]{robotNomenclature.pdf}}%
  \end{picture}%
\endgroup%